\documentclass[10pt]{wlscirep}
\usepackage[utf8]{inputenc}
\usepackage[T1]{fontenc}
\usepackage{booktabs,tabularx}
\usepackage{caption}
\usepackage{makecell} 
\usepackage{hyperref}
\title{Deep Learning Predicts Prevalent and Incident Parkinson’s Disease From UK Biobank Fundus Imaging}

\author[1]{Charlie Tran}
\author[1]{Kai Shen}
\author[2]{Kang Liu}
\author[3]{Akshay Ashok}
\author[4]{Adolfo Ramirez-Zamora}
\author[5]{Jinghua Chen}
\author[6]{Yulin Li}
\author[1,7,8,*]{Ruogu Fang}
\affil[1]{Department of Electrical and Computer Engineering, University of Florida, Gainesville, FL 32611, USA}
\affil[2]{Department of Physics, University of Florida, Gainesville, FL, 32611, USA}
\affil[3]{Dept. of Computer and Information Science and Engineering, University of Florida, Gainesville, FL. 32611, USA}
\affil[4]{Department of Neurology, University of Florida, Gainesville, FL 32611, USA}
\affil[5]{Department of Ophthalmology, University of Florida, Gainesville, FL 32611, USA}
\affil[6]{Department of Biostatistics, University of Florida, Gainesville FL, 32611, USA}
\affil[7]{J. Crayton Pruitt Family Department. of Biomedical Engineering, University of Florida, Gainesville, FL 32611, USA}
\affil[8]{Center for Cognitive Aging and Memory, University of Florida, Gainesville, FL 32611, USA}

\begin{document}

\flushbottom

\maketitle \\
*\textbf{Corresponding Author} \\ 
Ruogu Fang, Ph.D. \\
J. Crayton Pruitt Family Department of Biomedical Engineering \\
Herbert Wertheim College of Engineering \\ 
University of Florida \\ 
PO Box 116131, 1275 Center Drive \\ 
Gainesville, FL 32611-6131 \\ 
Phone: (352) 294-1375 \\
Email: ruogu.fang@ufl.edu

\section*{Abstract} 
Parkinson’s disease is the world’s fastest-growing neurological disorder. Research to elucidate the mechanisms of Parkinson’s disease and automate diagnostics would greatly improve the treatment of patients with Parkinson’s disease. Current diagnostic methods are expensive and have limited availability. Considering the insidious and preclinical onset and progression of the disease, a desirable screening should be diagnostically accurate even before the onset of symptoms to allow medical interventions. We highlight retinal fundus imaging, often termed a window to the brain, as a diagnostic screening modality for Parkinson’s disease. We conducted a systematic evaluation of conventional machine learning and deep learning techniques to classify Parkinson’s disease from UK Biobank fundus imaging. Our results show that Parkinson’s disease individuals can be differentiated from age and gender-matched healthy subjects with an Area Under the Curve (AUC) of 0.77. This accuracy is maintained when predicting either prevalent or incident Parkinson’s disease. Explainability and trustworthiness are enhanced by visual attribution maps of localized biomarkers and quantified metrics of model robustness to data perturbations.
\vskip5pt

\section*{Introduction}

Parkinson’s disease (PD) is one of the world’s fastest-growing neurological disorders, characterized by progressive impairment in motor control and multiple non-motor symptoms. \cite{a1, a2, a3} The manifestation of these symptoms is pathologically characterized by the significant loss of dopaminergic neurons in the substantia nigra. \cite{a4} An estimated one million individuals in the United States have PD, leading to nearly \$50 billion a year in economic burden. \cite{a1} Notably, this financial burden consists not only of direct medical costs but also indirect influences such as necessitated family care and social welfare. The World Health Organization estimates the prevalence of PD has doubled in the last 25 years, while the number of deaths caused by PD increased by over 100\% since 2000, largely due to the lack of effective intervention under the rising growth of the elderly population. \cite{a2} Innovations to our understanding of the pathology of PD and the development of early diagnostic systems are desired to address the global concerns arising from PD.

Systematic diagnostic evaluation of PD currently struggles due to the lack of early biomarkers and balance between both specificity and sensitivity. \cite{a3} Indeed, cardinal motor symptoms fall within the umbrella of Parkinsonism, while non-motor indicators are symptomatic of numerous neurodegenerative diseases. \cite{a5} Risk factors such as age, gender, and environmental toxin exposure are not specific to PD. Differential diagnosis frameworks established by the UK’s Parkinson’s Disease Society Brain Bank and the International Parkinson and Movement Disorder Society \cite{a6} are the current standards for evaluation. However, these \textit{checklists} require increasing amounts of exclusion and evidence, including response to dopaminergic therapy (levodopa), \cite{a7} occurrence of dyskinesia, \cite{a8} and even DaTscan imaging \cite{a9} to conclude a definitive PD diagnosis. A biological definition necessitating additional testing is being entertained. Moreover, early, or atypical PD complicates the observability of cardinal signs, leading to significantly reduced diagnostic accuracy. \cite{a10, a11} Thus, current diagnostic indicators lack solid predictive power, straining diagnostic expenses, time, availability, enrollment in therapeutic or disease-modifying trials, and subjectivity.

The retina provides a routine, inexpensive, and non-invasive modality for studying brain-related pathological processes of neurodegenerative diseases, often referred to as a \textit{window to the brain}.\cite{a12, a13, a14, a15} Dopamine plays a complex role in visual pathway processing, justifying visual dysfunction findings in PD individuals. \cite{a16} Moreover, substantial evidence has revealed large temporal gaps between the onset of PD and observable symptoms, inspiring the retina as a prodromal PD biomarker. \cite{a5} Clinical studies have suggested retinal layer thinning and reductions in microvasculature density in PD patients primarily through optical coherence tomography (OCT) and optical coherence tomography angiography (OCT-A).  \cite{a17, a18}  Nonetheless, clinical findings concerning both retinal degeneration and disease-specific information (disease duration, disease severity, etc.) are not always consistent, \cite{a19, a20} demanding further studies to bolster retinal diagnostic power.

Artificial intelligence (AI) algorithms are efficient diagnostic tools through their ability to identify, localize, and quantify pathological features, as evident from their success in diverse retinal disease tasks. \cite{a21, a22} In particular, supervised (labeled) algorithms employed in these tasks can generally be divided into two categories: conventional machine learning algorithms and deep learning models. Conventional machine learning models are known to create (non)linear decision boundaries by rules of inference. On the other hand, deep learning models have risen as powerful models due to their ability to learn meaningful representations and extract subtle features in the training process. Learning retinal biomarkers of PD demands an intricate understanding of structural degeneration of the retinal vasculature, a task unfeasible for even experienced ophthalmologists and neurologists. To address this challenge, we propose the usage of AI algorithms to extract the complex relationships existing within the global and local spatial levels of the retina.

We provide one of the first comprehensive artificial intelligence studies of PD classification from fundus imaging. Our key objective is realized by systematically profiling the classification performance across different stages of Parkinson’s disease progression, namely incident (consistent with pre-symptomatic and/or prodromal) PD and prevalent PD. Improving upon other related works, we maximize the diagnostic capacity of AI algorithms by neglecting the usage of any external quantitative measures or feature selection methods. Finally, we assess diagnostic consistency and robustness through extensive experimentation of both conventional machine learning and deep learning approaches together with a post-hoc spatial feature attribution analysis. Overall, this work enables future research in the development of efficient diagnostic technology and early disease intervention.

\section*{Results}

\subsection*{Study Design and Clinical Characteristics}

\begin{figure}[!h]
 \centering
 \includegraphics[width= 0.5\columnwidth]{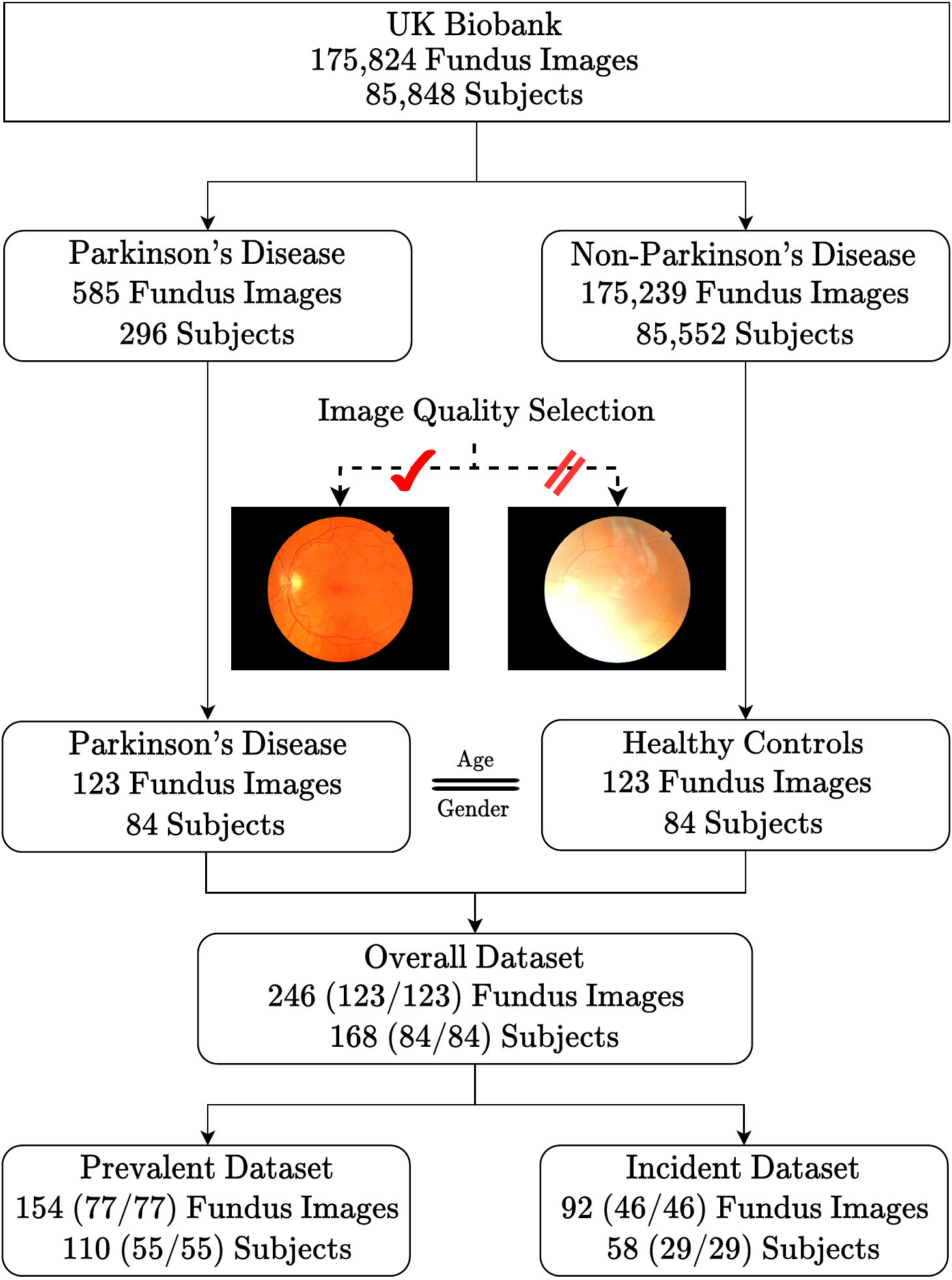} 
\caption{\textbf{Data Collection pipeline from the UK Biobank}. Instances in parentheses represent an equal balance of Parkinson’s disease and healthy controls subjects. Multiple quality selection phases were used as additional inclusion criteria into our dataset arising from AutoMorph and manual image grading. In total, we have the \textit{overall} dataset of PD subjects matched with age and gender-matched healthy controls, and two subsets corresponding to \textit{prevalent} and \textit{incident} subjects.}
 \label{fig: fig1}
\end{figure}

This study draws from the UK Biobank, a biomedical database containing over 500,000 participants aged 40-69 from 2006 to 2010. \cite{a23} As of 2019 October, 175,824 fundus images from 85,848 participants were available along with broad clinical health measures. Most subjects have two retinal photographs (left and right eye) with a minority having an additional follow-up imaging session. From this population, we identified 585 fundus images from 296 unique subjects. Following manual image quality selection guidelines, we determined 123 fundus images from 84 unique participants. To carry out an unbiased binary classification framework, we matched each PD image according to the subjects’ age and gender, that is, a healthy control cohort of 123 fundus images from 84 subjects. This constitutes our binary-labeled \textit{overall} dataset of 246 fundus images (123 PD, 123 HC) from 168 subjects (84 PD, 91 HC). Lastly, we form two subsets of the data corresponding to a \textit{prevalent} dataset of 154 fundus images (77 PD, 77 HC) from 110 subjects (55 PD, 55 HC) and an \textit{incident} dataset of 92 fundus images (46 PD, 46 HC) from 58 subjects (29 PD, 29 HC). Notably, we define the diagnostic gap as the difference between the date of image acquisition minus the date of diagnosis, where a negative value is interpreted as having a PD diagnosis before fundus image acquisition (prevalent PD) and a positive value is interpreted as having a PD diagnosis post fundus image acquisition. The data collection pipeline is summarized in Figure \ref{fig: fig1}. 

Risk factors of Parkinson's disease have been extensively studied, \cite{a24, a25, a26} including age, gender, ethnicity, Townsend deprivation indices, alcohol consumption, history of obesity-diabetes, history of stroke, and psychotropic medication usage. Moreover, the effects of Parkinson's Disease have been associated with visual symptoms, from which, we acquire a history of diagnostic eye problems and visual acuity measures. We detail the statistical analyses of subject demographics, visual measures, and covariates of our study population in Table \ref{table: table1}. \\

\subsection*{Model Design}

Conventional machine learning and deep learning models were systemically studied for their performance in PD diagnosis, a binary classification task. We evaluate the performance of each AI model by five randomized repetitions of stratified five-fold cross validation (at the subject level) based upon several classification metrics (AUC, accuracy, PPV, NPV, sensitivity, specificity, and F1-score). Given the insufficient amount of longitudinal data, we treat each image separately as a sample without data leakage, rather than performing a longitudinal analysis. The conventional machine learning models provide a classification performance baseline, from which we utilize Logistic Regression, Elastic-Net, Linear SVM and Radial Basis Function SVM kernels. On the other hand, we evaluate the performance of popular deep learning frameworks including AlexNet, VGG-16, GoogleNet, Inception-V3, and ResNet-50. We follow traditional guidelines such as image normalization to the training data for our machine learning models, ImageNet normalization, spatial augmentations, and early-stopping for our deep learning models. The detailed guidelines are outlined in Methods (``Data Pre-processing and Model Training”).

\subsection*{Model Performance in Overall PD Groups}

The best deep learning model was AlexNet with an average AUC of 0.77 (95\% CI, 0.74 – 0.81), accuracy of 0.68 (95\% CI, 0.65 – 0.72), PPV of 0.69 (95\% CI, 0.64 – 0.74), NPV of 0.79 (95\% CI, 0.72 – 0.86), sensitivity of 0.76 (95\% CI, 0.66 – 0.87), specificity of 0.60 (95\% CI, 0.51 – 0.70), and F1-score of 0.68 (95\% CI, 0.62 – 0.75). The best conventional machine learning model was the RBF support vector machine with an optimal AUC of 0.71 (95\% CI, 0.69 – 0.74), accuracy of 0.67 (95\% CI, 0.65 - 0.70), PPV of 0.65 (0.63 - 0.67), NPV of 0.71 (0.68 – 0.74), sensitivity of 0.76 (95\% CI, 0.72 – 0.80), specificity of 0.58 (95\% CI, 0.54 - 0.63), and F1-score of 0.70 (95\% CI, 0.68 – 0.72). The VGG-16, SVM (Linear), Logistic Regression, and ElasticNet perform similarly, while Inception-V3 and ResNet-50 perform significantly worse (Figure \ref{fig: fig2} and Table \ref{table: table2}). 

\subsection*{Model Performance in Prevalent PD Groups}

The best performing deep learning was the AlexNet model with an average AUC of 0.73 (95\% CI, 0.68 – 0.77),  accuracy of 0.65 (95\% CI, 0.62 – 0.68), PPV of 0.62 (95\% CI, 0.55 – 0.68), NPV of 0.69 (95\% CI, 0.61 – 0.77),  sensitivity of 0.74 (95\% CI, 0.65 – 0.83), specificity of 0.56 (95\% CI, 0.47 – 0.65), and F1-score of 0.66 (95\% CI, 0.60 – 0.72). The best performing conventional machine learning model was the SVM (Linear) with an AUC of 0.73 (95\% CI, 0.69 – 0.77), accuracy of 0.67 (95\% CI, 0.64 – 0.70), PPV of 0.67 (95\% CI, 0.63 – 0.72), NPV of 0.68 (95\% CI, 0.64 – 0.71), sensitivity of 0.69 (95\% CI, 0.64 – 0.73), specificity of 0.65 (95\% CI, 0.58 – 0.71), and F1-score of 0.67 (95\% CI, 0.64 – 0.70). Note that in contrast to the ``Overall PD Groups”, the SVM (Linear) outperformed the SVM (RBF).  In general, the performance of the artificial intelligence models was relatively slight upon the transition between “Overall” to “Prevalent” PD groups. Detailed comparisons can be found in Figure \ref{fig: fig2} and Table \ref{table: table2}. 

\subsection*{Model Performance in Incident PD Groups}

\begin{figure}[!t]%
\centering
\includegraphics[width=0.8\textwidth]{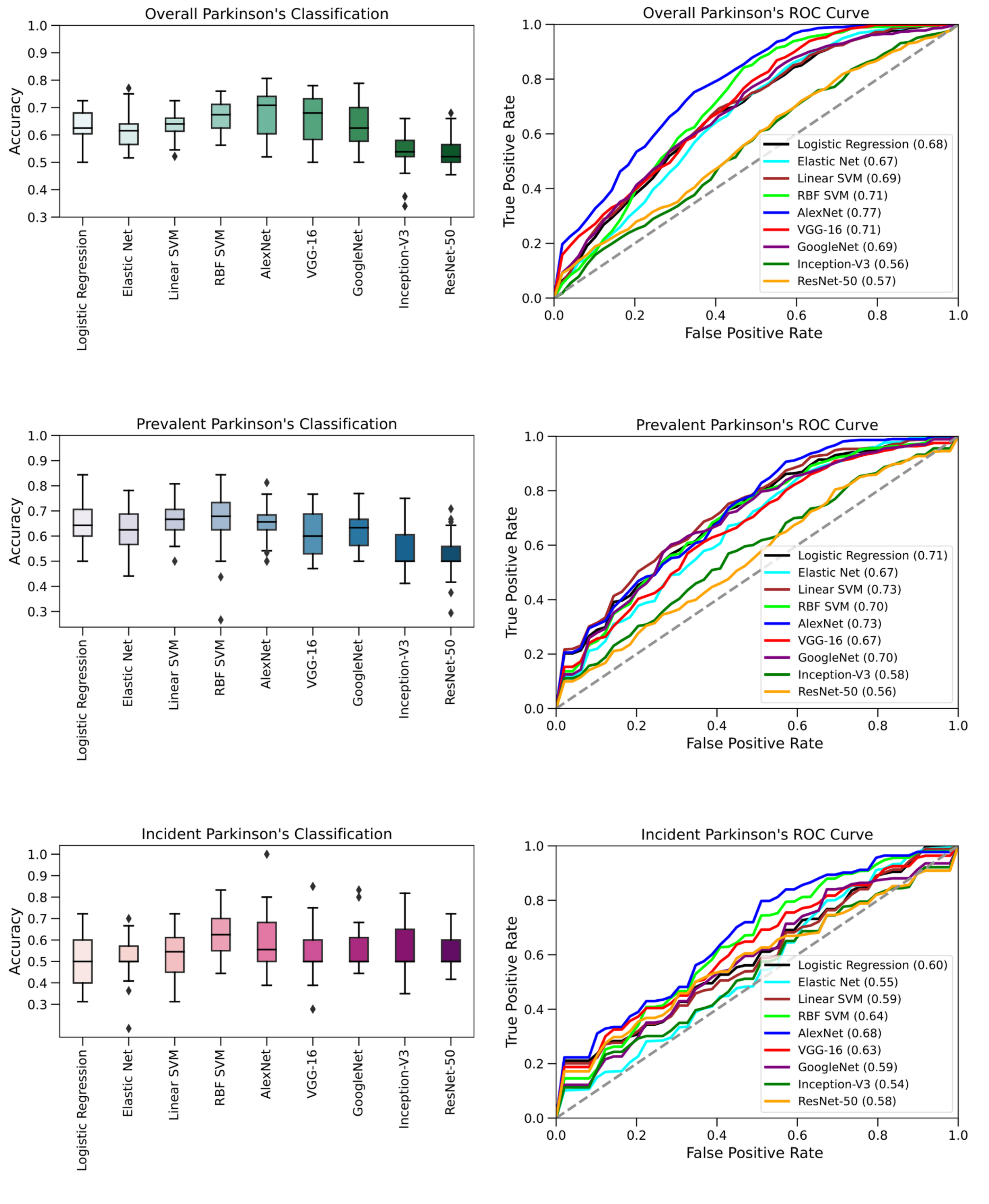}
\caption{\textbf{Box-plots and ROC curves of the Parkinson Disease Classification Models.} The models are evaluated over five randomized repetitions of the five-fold stratified cross-validation protocol. The AUC scores are enlisted in the legend.}
\label{fig: fig2}
\end{figure}

The best-performing deep learning model was the AlexNet model with an average AUC of 0.68 (95\% CI, 0.60 – 0.75), accuracy of 0.60\% (95\% CI, 0.54 – 0.66), PPV of 0.49 (95\% CI, 0.39 – 0.60), NPV of 0.66 (95\% CI, 0.56 – 0.77), sensitivity of 0.64 (95\% CI, 0.48 – 0.79), specificity of 0.56 (95\% CI, 0.45 – 0.68)  and F1-score of 0.54 (95\% CI, 0.42 – 0.67). The performance of the AlexNet and VGG-16 remains similar while other deep learning models decline with less than 65\% accuracy. The best-performing conventional machine learning model was the SVM (RBF) with an AUC of 0.64 (95\% CI, 0.59 – 0.70), accuracy of 0.62 (95\% CI, 0.58 – 0.67),  PPV of 0.63 (95\% CI, 0.58 – 0.68), NPV of 0.63 (95\% CI, 0.58 – 0.69), sensitivity of 0.64 (95\% CI, 0.59 – 0.69), specificity of 0.60 (95\% CI, 0.53– 0.67), and F1-score of 0.63 (95\% CI, 0.59 – 0.67). In general, the performance of the machine learning classifiers decreased significantly relative to the overall and prevalent groups. A summarized completion of the results can be found in Figure \ref{fig: fig2} and Table \ref{table: table2}.

\subsection*{Visualization and Explainability}
\begin{figure}[!t]%
\centering
\includegraphics[width=0.76\linewidth]{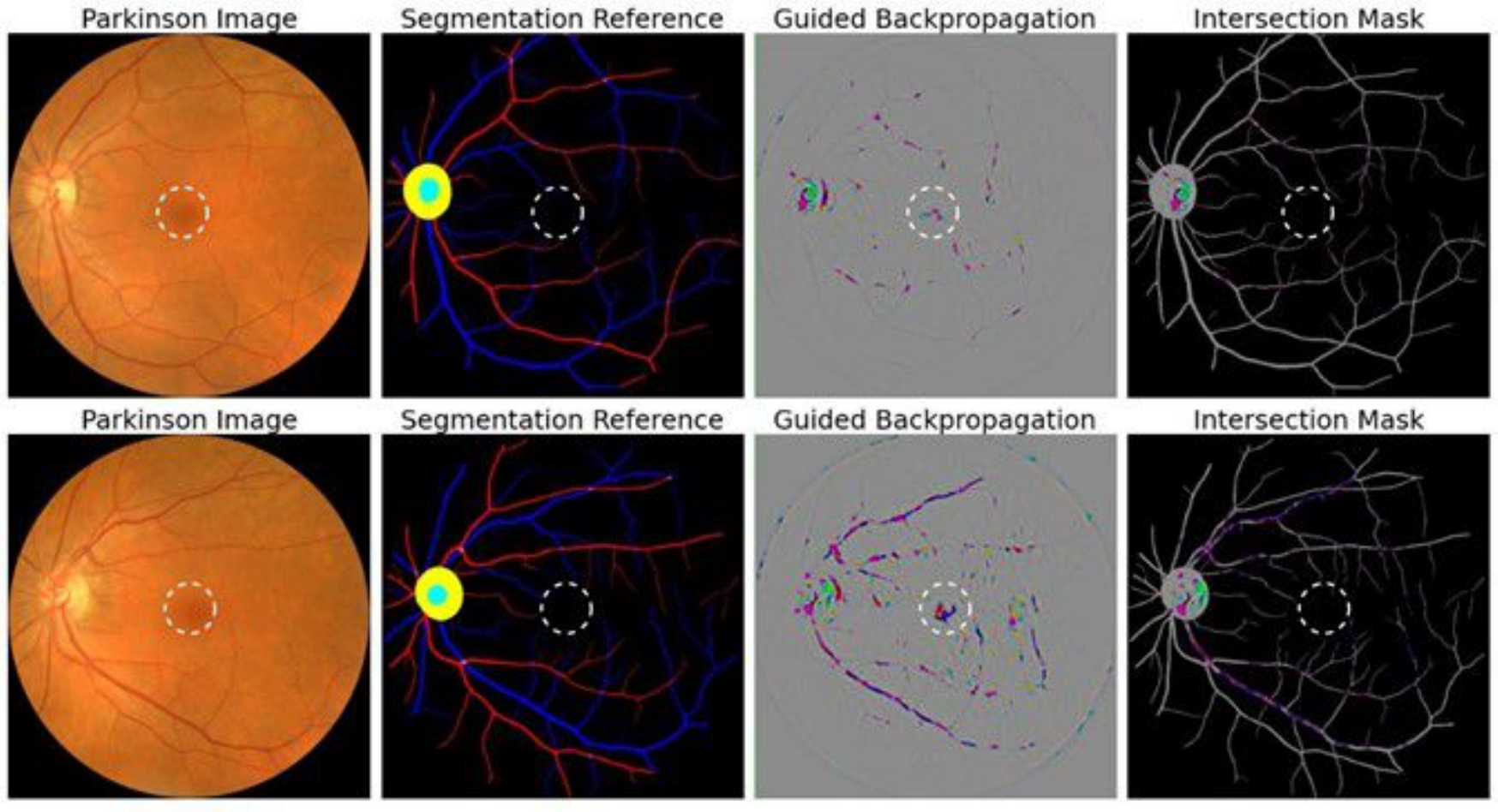}
\caption{\textbf{Attribution correspondence of retinal features}. In the first column, an artery-vein (red and blue, respectively) map is combined with the optic cup (teal) and optic disc (yellow) generated from the AutoMorph deep learning segmentation module. A white dashed line is shown as an estimate for the foveal region. In the third column, a predicted attribution map is generated using the guided backpropagation algorithm on top of the AlexNet model. The intersection of the salient features with the segmentation is shown in the last column. The images represent the left (top) and right (bottom) eyes from the same subject, demonstrating distinct feature distributions for prediction.   }
\label{fig: fig3}
\end{figure}

\begin{figure}[!h]%
\centering
\includegraphics[width=0.8\linewidth]{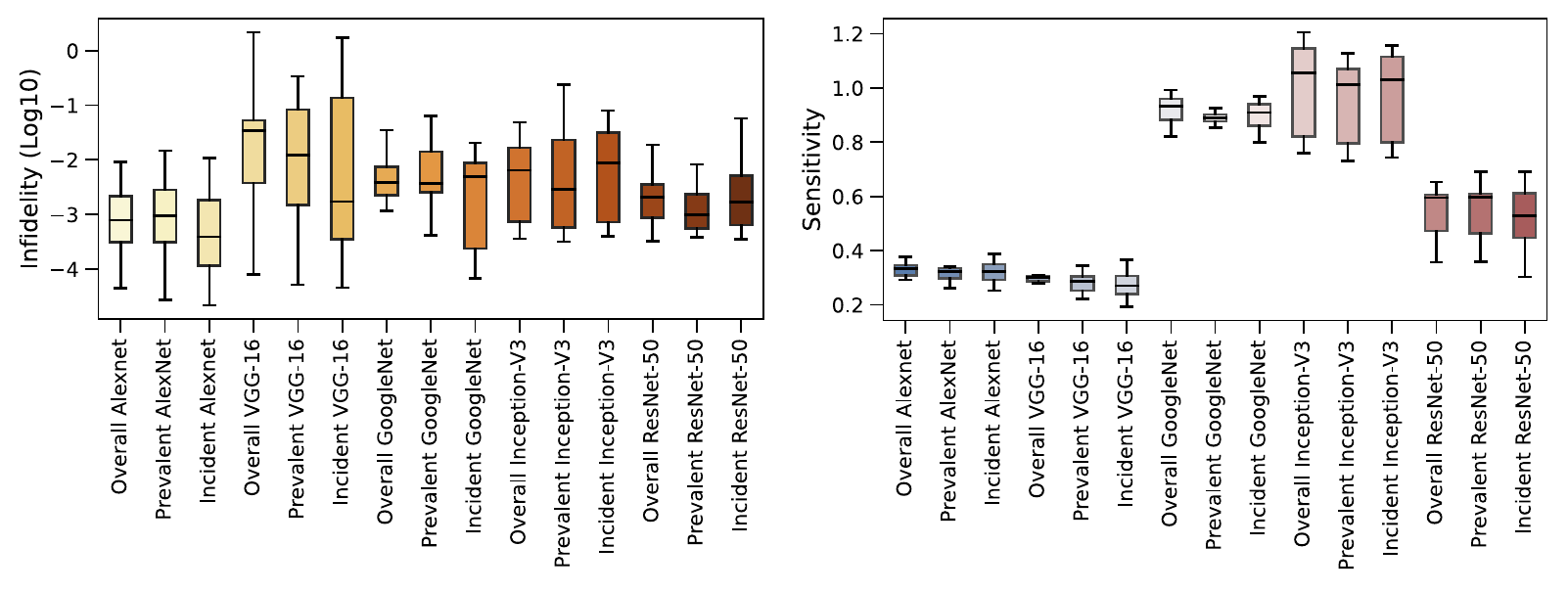}
\caption{\textbf{Explanation of infidelity and sensitivity comparison} among different models. The logarithm of the infidelity score was applied for visualization due to the large range of scores. }
\label{fig: fig4}
\end{figure}

Qualitative explainability is visualized through guided backpropagation on our deep learning models, revealing the models were able to correctly distinguish subtle features on the retinal anatomy (Figure \ref{fig: fig3}). To highlight that the extracted features are consistent with those recognized as retinal biomarkers of neurodegeneration in PD, we use the AutoMorph\cite{a27} deep learning segmentation module to generate a map of important retinal structures, namely the arteries, veins, optic cup, and optic disc, along with a manually marked fovea. The optic cup, optic disc, and fovea have been shown to flag progression in PD via observable structural changes in size, thickness, and other features.\cite{a28, a29, a30} As such, the overlay of the guided backpropagation map and segmentation map yields a visual comparison of the consistency between features used for classification and potential retinal biomarkers of PD. Moreover, we reinforce the qualitative explainability by quantifying the robustness of our deep learning models to data perturbations through an infidelity and sensitivity analysis. The definitions of the infidelity and sensitivity measures are outlined in Method (``Explainability Evaluation''). In this step, we run the guided backpropagation attribution step across N = 50 perturbations at test time with a noise distribution drawn from a normal distribution $N \sim (0,0.01^2)$. The results of the infidelity and sensitivity analysis are averaged across the repeated cross validation protocol and summarized in Figure \ref{fig: fig4}. Our empirical results suggest that AlexNet is the most robust to data perturbations, while simultaneously the most accurate according to lower infidelity and sensitivity measures. Notably, the VGG-16 has similar robustness measurers as AlexNet with less classification performance, possibly owing to the large model parameter complexity. Predictions made by other deep learning architectures including GoogleNet, Inception-V3, and ResNet-50 were largely influenced by data perturbations.

\subsection*{Feature Engineering Analysis}
Feature engineering approaches are generally devised to enhance AI outcomes, in particular, conventional machine learning models, due to the large feature space complexity. On the other hand, deep learning models by design can achieve performance with minimal amounts of data pre-preprocessing. We explore the performance of feature engineering in conventional machine learning models with two changes to the input fundus images: (1) gray-scale color conversion (reducing the dependence of color), and (2) vessel-segmentation via AutoMorph (emphasizing the retinal vasculature). The approach adopted in the latter is also a useful comparison with that of Tian et al. which utilized the technique for Alzheimer’s Disease detection. The results of this approach are demonstrated in Supplementary Table \ref{table: S4}. Conventional machine learning models are demonstrated to improve by simple color conversion, while declining using the retinal vasculature. The reduction in performance using the vessel segmentation algorithm hints that essential diagnostic features exist across different regions of the eye (e.g. the optic cup and fovea).

\subsection*{Model Covariate Analysis - Diagnostic Gap and Gender}

We investigate the influence of Parkinson’s disease progression on the model performance. The Parkinson’s disease progression can be expressed by the diagnostic gap and thereby a proxy measure of disease severity. Treating each image independently, we divided our Parkinson’s subjects into four quartiles according to the diagnostic time (years) and examined the model performance compiled over our repeated five-fold cross validation in our best model, AlexNet. Sensitivity measures in the Parkinson’s group did not exhibit monotonic relationships based on the diagnostic gap or consistencies in dataset subtypes (see Supplementary Table \ref{table: S1}). One such explanation for this result is discrepancies in the diagnostic gap, that is, the dates of diagnosis do not co-align with the true dates of disease acquisition as Parkinson’s disease is a progressive disease. Otherwise, a claim could be made that the model performance is not contingent strictly on the diagnostic gap, but rather, an individual basis of existing imaging biomarkers.  

Gender is a known risk factor of Parkinson’s disease.\cite{a31,a32} We investigated a potential bias in our deep learning models due to gender differences in the retina. We categorized the model performances of our AlexNet model compiled over all subjects, Parkinson’s-specific, and healthy-control specific data, and conducted a series of Chi-Square Test of Independence (see Supplementary Table \ref{table: S2}). In cases where the frequency of observations was found to be less than or equal to five, we conducted a Fisher’s Exact test. We discover no statistical significance (p < 0.05) for any experiment or data-subtype.

\section*{Discussion}
This work demonstrates that deep neural networks can be trained to detect Parkinson’s disease in retinal fundus images with decent performance. Our model can predict the incidence of Parkinson’s disease ahead of formal diagnosis at demonstrated sensitivity levels of 80.0\% from 0 to 3.93 years, 80.0\% from 3.93 to 5.07 years, 93.33\% from 5.07 to 5.57 years, and 81.67\% from 5.57 to 7.38 years. These results indicate a potential pathway for early disease intervention. Automated deep neural networks show strong promises to assist and complement ophthalmologists in terms of biomarker identification and high-throughput evaluation.

Artificial intelligence evaluation of Parkinson’s disease through the retina has been rarely applied. Hu et al.\cite{a33} trained a deep learning model to evaluate the retinal age gap as one predictive marker for incident Parkinson’s Disease using fundus images from the UK Biobank, showcasing statistical significance and predictive AUC of 0.71. Nunes et al.\cite{a34} used optical coherence tomography data to compute retinal texture markers and trained a deep learning model with a median sensitivity of 88.7\%, 79.5\%, and 77.8\% concerning healthy controls, Parkinson’s disease, and Alzheimer’s disease. However, all these works did not provide a comprehensive comparison of conventional machine learning and deep learning methods on this problem and lacked insights into the explainability of their models. We extend upon these works by treating the entire fundus image as a diagnostic modality, and comprehensively evaluate a broad spectrum of conventional machine learning and deep learning methods, as well as shedding light on the explainability in both image space and on the algorithm level. Our work will lay a solid foundation for future exploration in this direction and serve as a reference for algorithm selection in terms of both performance and explainability. Related works have been accomplished focusing specifically on Alzheimer’s disease, e.g., Tian et. al\cite{a35} and Wisely et al.\cite{a36} but not Parkinson’s disease. Clinical studies in the field have yielded statistical differences in the retinal layers between PD and HC subjects, lacking evidence for diagnostic power. Further work is necessitated in the field of deep learning to build stronger classification performance and understanding of retinal biomarkers. In the future, a multi-modal model utilizing optical coherence tomography, fundus autofluorescence, and/or electronic health records is a considerable direction for Parkinson’s disease analysis.

This study has some limitations. First, the size of our dataset could be enlarged to further capture the wide presentations of Parkinson’s disease. Moreover, the data is derived from the UK population and therefore future studies are needed to evaluate whether these models can generalize to other populations. So far, public datasets containing both Parkinson’s disease subjects and fundus images (as well as patient health records) apart from the UK Biobank are not available, thus new datasets including both retinal images and PD diagnosis in a larger population will be helpful. A deeper study would hope to investigate the severity of Parkinson’s disease, e.g. by the MDS-Unified Parkinson's Disease Rating Scale (MDS-UPDRS), wherein the severity was (weakly) substituted by the diagnostic gap in this work. Furthermore, this current research has been restricted to Parkinson’s disease, and it remains questionable whether different eye diseases or neurogenerative diseases (e.g., Alzheimer’s Disease) share identical biomarkers or degeneration patterns. The next major research question is whether such model explanations are consistent and/or able to guide the grading of ophthalmologists, which is a major goal of clinical translational research. This matter is further complicated as the visual biomarkers for Parkinson’s disease are less well-understood than common eye diseases such as glaucoma. Prospective assessments of retina imaging coupled with biological details and clinical phenotyping are needed to provide insight into the use and implementation of these techniques. These limitations necessitate future work to ensure the trustworthiness of artificial intelligence models in a clinical setting.

Deep learning models outperformed conventional machine learning models to accurately predict Parkinson’s disease from retinal fundus images. We demonstrate deep learning models can nearly equally diagnose both prevalent and incident PD subjects with robustness to image perturbations, paving the way for early treatment and intervention. Further studies are warranted to verify the consistency of Parkinson’s disease evaluation, to enhance our understanding of retinal biomarkers, and to incorporate automated models into clinical settings.
 
\section*{Methods}

\subsection*{UK Biobank Participants}
The UK Biobank (UKB) is one of the largest biomedical databases, recruiting over 500,000 individuals aged 40-69 years old at baseline throughout assessment centers in the United Kingdom in 2006-2010. The methods by which this data was acquired have been described elsewhere.\cite{a23} Diverse patient health records include demographic, genetic, lifestyle, and health information. Comprehensive physical examinations as well as ophthalmic examinations were conducted for further analysis. Health-related events were determined using data linkage to the Health Episode Statistics (HES), Scottish Morbidity Record (SMR01), Patient Episode Database for Wales (PEDW), and death registers.

\subsection*{Parkinson's Disease and Definitions} 
Parkinson’s disease was determined by hospital administration data in the United Kingdom, national death register data, and self-reported data. We consider prevalent PD subjects diagnosed prior to baseline assessment and incident PD subjects diagnosed following baseline assessment. Prevalent PD subjects were labeled according to hospital admission electronic health records based on the International Classification of Diseases (ICD9, ICD10) codes or self-reports. Incident PD subjects were labeled according to either the ICD codes or the death registry. The earliest recorded diagnostic dates take priority in case of multiple records. If PD was recorded in the death register only (diagnosed post-mortem), the date of death is used for the date of diagnosis. We define the diagnostic gap as the difference between the date of image acquisition minus the date of diagnosis, where a negative value is interpreted as having a PD diagnosis prior to fundus image acquisition (prevalent PD), and a positive value is interpreted as having a PD diagnosis post fundus image acquisition (incident PD). We acquire our PD labels according to the UKB Field 42032. Further details may be inquired upon from the UKB documentation.\cite{a37}

\subsection*{Ophthalmic Measures}
In the UKB eye and vision consortium, the ophthalmic assessment included (1) questionnaires of past ophthalmic and family history, (2) quantitative measures of visual acuity, refractive error, and keratometry, and (3) imaging acquisition including spectral domain optical coherence tomography (SD-OCT) of the macula and a disc-macula fundus photograph. In our study, we acquire diagnostic fields of eye problems (glaucoma, cataracts, diabetes-related, injury/trauma, macular, degeneration, etc.), visual acuity measured as the logarithm of the minimum angle of resolution (LogMar), and fundus photographs. Fundus photographs were acquired using a Topcon 3D OCT-1000 Mark II system. The system has a $45^{\circ}$ field angle, scanning range of 6 mm $\times$ 6 mm centered on the fovea, acquisition speed of 18,000 A-scans per second, and 6 µm axial resolution. The details of the eye and vision consortium have been described in other studies.\cite{a38}

\subsection*{Study Population and Summary Statistics}
A total of 175,824 fundus images from 85,848 subjects were discovered in the UKB (as of 2019 October, UKB). Among this population, we found 585 fundus images from 296 subjects with PD. Image quality selection was held in multiple phases: (1) the deep learning image quality selection module of AutoMorph\cite{a27} pretrained on Eye-PACS-Q\cite{a39}, (2) subjectivity to the quality of the vessel and optic cup and disc AutoMorph segmentation, and (3) manual grading borderline images according to external guidelines,21 including artifacts, clarity, and field definition defects. Each phase is held in sequence with the additional rounds held to justify borderline candidates for inclusion or exclusion into the dataset. In these guidelines, an image quality score is determined from the total of artifacts (-10 to 0), clarity (0 to +10), and field definition (0 to +10) where an optimal score is +20 and a score less than or equal to 12 is rejectable. The artifacts component evaluates the broad proportion of visibility incurred by artifacts in the image, clarity evaluates the relative visibility of veins and lesions in the image, and field definition evaluates the broad field of view of key retinal structures (e.g. the optic cup and disc, and fovea). Of note, three images categorized as ungradable by AutoMorph, were moved into our dataset on the basis of sufficient manual grading and viewing of the retinal vasculature. Following manual selection, a total of 123 usable Parkinson’s disease images from 84 subjects were found to have met our criteria for inclusion. For each PD image, a healthy control (HC) with no history of PD was matched according to their age and gender to prevent covariate bias using the aforementioned image quality selection guidelines. All other fundus images and corresponding subjects were excluded. This constitutes our binary-labeled overall dataset of 246 fundus images (123 PD, 123 HC) from 168 subjects (84 PD, 84 HC). Lastly, we form two subsets of the data corresponding to a prevalent dataset of 154 fundus images (77 PD, 77 HC) from 110 subjects (55 PD, 55 HC) and an incident dataset of 92 fundus images (46 PD, 46 HC) from 58 subjects (29 PD, 29 HC).

\subsection*{Data-Preprocessing and Model Training}

This study explores both deep learning and conventional machine learning models. Specifically, our deep learning models are convolutional neural networks (CNN) including AlexNet,\cite{a40} VGG-16,\cite{a41} GoogleNet,\cite{a42} Inception-V3,\cite{a43} and ResNet-50.\cite{a44}. Our conventional machine learning models include Logistic Regression, Elastic Net (regularized with hyperparameters $\lambda_{L_{1}} = 1$, $\lambda_{L_{2}} = 0.5$, where L1 and L2 designate the respective norms), and support vector machines (linear and radial basis function kernels). 

Our deep learning models are initialized with ImageNet classification weights, with the input data normalized to the mean and standard deviation of ImageNet. The input of our conventional machine learning models is standardized to unit variance according to the training set’s mean and standard deviation. Our testing evaluation is designed using a five-fold stratified cross-validation, wherein each Parkinson’s subject is exactly matched to its corresponding age and gender matched control to remove bias of age and gender within the training and evaluation of our models. To further emphasize our training procedure, one-fold is selected as a test set, while the remaining four folds are delegated for allowable training. Of the four folds, one is selected as an internal validation set for hyper-parameter tuning while the model is initially trained using three of the training folds. For our conventional machine learning models, the regularization C-parameter in the set [0.1, 0.01, 1, 10, 100] is optimized. On the other hand, we fine-tune our deep learning models pre-trained on ImageNet using a binary cross-entropy loss, Adam optimizer,\cite{a45} learning rate of $1e-4$, batch size of 64, and 100 epochs with early stopping. The choice of the learning rate was chosen to be roughly optimal and stable upon early experimentation of learning rates ranging from $1e-1$ to $1e-4$. After the discovery of the optimal hyper-parameters, the model re-consolidates the training set as the collection of all four training folds, and evaluated on the final test-set.

\subsection*{Performance Evaluation}
The performance of our PD classifiers is averaged over five randomized repetitions of five-fold stratified cross-validation, for a total of 25 testing evaluations. We consider the area under the receiver operating characteristic curve (AUC), accuracy, positive predictive value (PPV), negative predictive value (NPV), sensitivity (true positive rate), specificity (true negative rate), and the F1-score. Due to the lack of longitudinal data for the same subject, the metrics are evaluated on a per-image basis, treating each sample independently without data leakage (subject-level cross validation-level split).  For reference, the training time, testing time, and number of model parameters are recorded (Supplementary Table \ref{table: S4}). 

\subsection*{Explainability Evaluation}

Qualitative visual explanations for our deep learning model predictions are accomplished by the Guided Backpropagation algorithm \cite{a46}, allowing an interpretable heat map of significant features. Quantitative explanations are provided by the explanation infidelity (INFD) and explanation sensitivity (SENS) metrics (to be distinguished from the classification performance metric sensitivity), where a lower result for both INFD and SENS provides evidence for better-robust models. The explanation infidelity and explanation sensitivity are computed at test-time over 50 perturbations drawn from a probability distribution $N \sim (0,0.01^2)$, the definitions of which are shown below, and the details discussed in the corresponding paper.\cite{a47}
\[ \text{SENS}_{\text{MAX}} (\Phi, \mathbf{f}, \mathbf{x}, r) = \max_{||\mathbf{y} - \mathbf{x}|| \leq r } || \Phi (\mathbf{f}, \mathbf{y} ) - \Phi (\mathbf{f}, \mathbf{x} ) || \] 
\[ \text{INFD}(\Phi, \mathbf{f}, \mathbf{x} ) = \mathbb{E}_{\mathbf{I}}\left [ \left ( \mathbf{I}^{T} \Phi (\mathbf{f}, \mathbf{x} ) - (\mathbf{f} (\mathbf{x} ) - \mathbf{f} (\mathbf{x} - \mathbf{I} ) ) \right )^2 \right ] \] 


\noindent where $\Phi$ is an explanation function (e.g., Guided Backpropagation), $\mathbf{f}$ is a black-box model (e.g., CNN), $\mathbf{x}$ is in the input (e.g., fundus image), $r$ is the input neighborhood radius, and $\mathbf{I}$ is the perturbation, here drawn from the noise distribution $\mathcal{N} \sim (0, 0.01^2)$. \\


\noindent \textbf{Acknowledgements} This study was supported by NSF IIS 2123809. \\

\noindent \textbf{Data Availability Statement:} This research has been conducted using the UK Biobank, a publicly accessible database, under Application Number 48388. The datasets are available to researchers through an open application via \url{https://www.ukbiobank.ac.uk/register-apply/}.  \\

\noindent \textbf{Code Availability Statement:} The underlying code for this study may be given permission by the authors from the GitHub repository \url{https://github.com/lab-smile/RetinaPD} but permission to the data is restricted to UKB applicants. \\

\noindent \textbf{Conflict of Interest Disclosures:} The authors declare no competing interests.  \\

\noindent \textbf{Author contributions:} Dr. Ruogu Fang and Charlie Tran had full access to all of the data in the study and take responsibility for the integrity of the data and the accuracy of the data analysis. 

\begin{itemize}[label={}, leftmargin=*, itemsep=0pt]
    \item \textit{Study concept and Design:} Charlie Tran, Kai Shen, Ruogu Fang.
    \item \textit{Acquisition, analysis, or interpretation of data:} Charlie Tran, Kai Shen, Kevin Liu, Ruogu Fang.
    \item \textit{Drafting of the manuscript:} Charlie Tran, Kai Shen, Kevin Liu, Yulin Li, Ruogu Fang.
    \item \textit{Critical revision of the manuscript for important intellectual content:} Charlie Tran, Ruogu Fang. Adolfo Ramirez-Zamora, Jinghua Chen, Akshay Ashok.
    \item \textit{Statistical analysis:} Charlie Tran, Kai Shen, Yulin Li.
    \item \textit{Obtained funding:} Ruogu Fang.
    \item \textit{Administrative, technical, or material support:} Ruogu Fang.
    \item \textit{Study Supervision:} Ruogu Fang.
\end{itemize}

\clearpage

\bibliography{sample.bib}

\newpage

\begin{table}[h!] 
\centering
\resizebox{\textwidth}{!}{
\begin{tabular}{l | l l l | l l l } 
\toprule
\textbf{Baseline Characteristics} & \textbf{PD Group} & \textbf{HC Group} &  \textbf{p-value} & \textbf{Prevalent} & \textbf{Incident} & \textbf{p-value} \\ 
\hline
Diagnostic gap, mean (SD), years & & & & -5.7 (4.5) & 4.8 (1.4) & -- \\
\hline
N & 84 & 84 & -- & 55 & 29 & -- \\ 
\hline
Age, mean (SD), years  &  61.8 (5.9)   & 61.8 (5.9) & 0.97 & 61.4 (5.8) & 62.7 (6.1)  &  0.33
\\ 
\hline
Gender, No. (\%) & & & 1.0 & & & 0.97\\
\hspace{3mm} Male & 49 (58.3) & 49 (58.3) &   &  32 (58.2) & 17 (58.6) &  \\
\hspace{3mm} Female & 35 (41.7)  &  35 (41.7) &  & 23 (41.8)  & 12 (41.4) &   \\

\hline
Ethnicity, No. (\%) &  &  & 0.56  & & & 0.64 \\
\hspace{3mm} White & 82 (97.6)  & 83 (98.2) & & 54 (98.2) & 28 (96.6) & \\
\hspace{3mm} Others  & 2 (2.4) & 1 (1.2) & & 1 (1.8) &  1 (3.4) & \\ 
\hline
Eye Problems, No. (\%) &  &  & 0.15 & & & 0.94 \\
\hspace{3mm} Yes & 9 (10.7)  & 4.0 (4.8) &  & 6 (10.9) & 3 (10.3) &  \\
\hspace{3mm} No & 75 (89.3) & 80 (95.2)  &   & 49 (89.1) & 26 (89.7) &  \\
\hspace{8mm} Diabetes-Related & 0 & 1 & {} & 0 & 0 & {} \\ 
\hspace{8mm} Glaucoma & 2 & 0 & {} & 0 & 2 & {} \\ 
\hspace{8mm} Injury/Trauma & 1 & 0 & {} & 1 & 0 & {} \\ 
\hspace{8mm} Cataracts & 5 & 2 & {} & 4 & 1 & {} \\ 
\hspace{8mm} Macular Degeneration & 0 & 1 & {} & 0 & 1 & {} \\ 
\hspace{8mm} Other & 1 & 0 & {} & 1 & 0 & {} \\

\hline
Visual Acuity, mean (SD)  & 0.28 (0.38)  & 0.04 (0.37) & * & 0.28 (0.37) & 0.29 (0.41) & 0.92  \\
\hline
Townsend Index, mean (SD) & -1.48 (2.95) & -2.68 (2.17) & 0.73 & -1.59 (3.09) & -1.27 (2.68) & 0.64\\
\hline
Smoking Status, No. (\%) & & & 0.62 & & & 0.08 \\
\hspace{3mm} Former/Current & 30 (35.7)  & 27 (32.1)  &  & 16 (29.1) & 14 (48.3)   \\ 
\hspace{3mm} Non-Smoker & 54 (64.3)  & 57 (67.9)  &  & 39 (70.9) & 15 (51.7)  \\
\hline
Drinking Status, No. (\%) & & & 0.63 & & & 0.61 \\ 
\hspace{3mm} Former/Current & 80 (95.2) & 84 (92.3) &  & 50 (94.3) & 30 (97.8) \\ 
\hspace{3mm} Non-Smoker & 4 (4.8) &  7 (7.7) &  & 3 (5.7) & 1 (3.2) \\
\hline
Obesity-Diabetes, No. (\%) & & & 0.46 & & & 0.14 \\ 
\hspace{3mm} Yes & 21 (25.0)  &  17 (20.2) &  & 11 (20.0) & 10 (34.5) \\ 
\hspace{3mm} No & 63 (75.0) & 67 (79.8) &  & 44 (80.0) & 19 (65.5)  \\
\hline
History of Stroke, No. (\%) & & & 0.47 & & & 0.23 \\ 
\hspace{3mm} Yes & 3 (3.6) & 5 (6.0) &  &  1 (1.8) & 2 (6.9) \\ 
\hspace{3mm} No & 81 (96.4)& 79 (94.0) &  & 54 (98.2) & 27 (93.1)   \\
\hline
Psychotropic Medication, No. (\%) & & & 0.32 & & & --  \\ 
\hspace{3mm} Yes & 0 (0) & 1 (1.2) &  & 0 (0) & 0 (0) \\ 
\hspace{3mm} No & 84 (100) & 83 (98.8) &  & 55 (100) & 29 (100) \\ 
\bottomrule
\end{tabular}}
\caption{\textbf{Baseline Characteristics of the study populations}. P-values conducted on continuous data are computed by the Student's t-test while categorical (binary) variables are computed by Pearson's Chi-squared test. * indicates statistically significant ($p < 0.05$). }
\label{table: table1}
\end{table}

\newpage
\begin{table}[htbp]
\centering
\resizebox{\textwidth}{!}{%
\begin{tabular}{@{}lccccccc@{}}
\toprule
\textbf{Model} & \textbf{AUC} & \textbf{ACC} & \textbf{PPV} & \textbf{NPV} & \textbf{SENS} & \textbf{SPEC} & \textbf{F1} \\ \midrule
\multicolumn{8}{c}{Overall (n = 246 fundus images | 123 PD, 123 HC)} \\ \midrule
Logistic Regression & 0.68 (0.65, 0.71) & 0.63 (0.61, 0.66) & 0.63 (0.6, 0.66) & 0.64 (0.61, 0.67) & 0.66 (0.62, 0.70) & 0.61 (0.55, 0.66) & 0.64 (0.62, 0.66) \\
Elastic Net & 0.67 (0.64, 0.69) & 0.61 (0.58, 0.64) & 0.60 (0.57, 0.62) & 0.63 (0.60, 0.67) & 0.68 (0.64, 0.72) & 0.54 (0.49, 0.58) & 0.64 (0.61, 0.66) \\
SVM (Linear) & 0.69 (0.66, 0.71) & 0.63 (0.61, 0.66) & 0.64 (0.61, 0.66) & 0.64 (0.62, 0.66) & 0.65 (0.61, 0.68) & 0.62 (0.57, 0.67) & 0.64 (0.61, 0.66) \\
SVM (RBF) & \textcolor{blue}{0.71 (0.69, 0.74)} & \textcolor{blue}{0.67 (0.65, 0.70)} & \textcolor{blue}{0.65 (0.63, 0.67)} & \textcolor{blue}{0.71 (0.68, 0.74)} & \textcolor{blue}{0.76 (0.72, 0.80)} & \textcolor{blue}{0.58 (0.54, 0.63)} & \textcolor{blue}{0.70 (0.68, 0.72)} \\
Alexnet & \textcolor{red}{0.77 (0.74, 0.81)} & \textcolor{red}{0.68 (0.65, 0.72)} & \textcolor{red}{0.69 (0.64, 0.74)} & \textcolor{red}{0.79 (0.72, 0.86)} & \textcolor{red}{0.76 (0.66, 0.87)} & \textcolor{red}{0.60 (0.51, 0.7)} & \textcolor{red}{0.68 (0.62, 0.75)} \\
VGG-16 & 0.71 (0.67, 0.75) & 0.66 (0.62, 0.69) & 0.61 (0.58, 0.64) & 0.81 (0.71, 0.91) & 0.85 (0.76, 0.95) & 0.46 (0.37, 0.54) & 0.70 (0.64, 0.76) \\
GoogleNet & 0.69 (0.65, 0.73) & 0.63 (0.60, 0.66) & 0.62 (0.58, 0.65) & 0.70 (0.62, 0.79) & 0.77 (0.69, 0.85) & 0.49 (0.41, 0.58) & 0.67 (0.63, 0.71) \\
Inception-V3 & 0.56 (0.52, 0.60) & 0.54 (0.51, 0.57) & 0.54 (0.49, 0.60) & 0.59 (0.52, 0.66) & 0.68 (0.57, 0.80) & 0.40 (0.29, 0.50) & 0.57 (0.50, 0.64) \\
ResNet-50 & 0.57 (0.53, 0.62) & 0.54 (0.51, 0.57) & 0.52 (0.40, 0.63) & 0.55 (0.45, 0.64) & 0.52 (0.37, 0.68) & 0.55 (0.41, 0.70) & 0.45 (0.34, 0.56) \\ \midrule
\multicolumn{8}{c}{Prevalent (n = 146 fundus images | 73 PD, 73 HC)} \\ \midrule
Logistic Regression & 0.71 (0.67, 0.76) & 0.65 (0.62, 0.69) & 0.66 (0.61, 0.70) & 0.66 (0.63, 0.70) & 0.68 (0.63, 0.73) & 0.63 (0.56, 0.69) & 0.66 (0.62, 0.70) \\
Elastic Net & 0.67 (0.62, 0.71) & 0.62 (0.59, 0.66) & 0.62 (0.57, 0.66) & 0.63 (0.56, 0.70) & 0.71 (0.65, 0.78) & 0.53 (0.45, 0.61) & 0.65 (0.61, 0.69) \\
SVM (Linear) & \textcolor{blue}{0.73 (0.69, 0.77)} & \textcolor{blue}{0.67 (0.64, 0.70)} & \textcolor{blue}{0.67 (0.63, 0.72)} & \textcolor{blue}{0.68 (0.64, 0.71)} & \textcolor{blue}{0.69 (0.64, 0.73)} & \textcolor{blue}{0.65 (0.58, 0.71)} & \textcolor{blue}{0.67 (0.64, 0.70)} \\
SVM (RBF) & 0.70 (0.64, 0.76) & 0.66 (0.60, 0.71) & 0.64 (0.57, 0.71) & 0.69 (0.63, 0.75) & 0.70 (0.61, 0.78) & 0.62 (0.55, 0.68) & 0.66 (0.59, 0.73) \\
Alexnet & \textcolor{red}{0.73 (0.68, 0.77)} & \textcolor{red}{0.65 (0.62, 0.68)} & \textcolor{red}{0.62 (0.55, 0.68)} & \textcolor{red}{0.69 (0.61, 0.77)} & \textcolor{red}{0.74 (0.65, 0.83)} & \textcolor{red}{0.56 (0.47, 0.65)} & \textcolor{red}{0.66 (0.60, 0.72)} \\
VGG-16 & 0.67 (0.61, 0.73) & 0.61 (0.57, 0.65) & 0.58 (0.51, 0.64) & 0.63 (0.55, 0.71) & 0.66 (0.55, 0.76) & 0.56 (0.47, 0.65) & 0.60 (0.52, 0.67) \\
GoogleNet & 0.70 (0.66, 0.74) & 0.62 (0.59, 0.66) & 0.60 (0.54, 0.66) & 0.68 (0.63, 0.74) & 0.69 (0.60, 0.79) & 0.55 (0.46, 0.65) & 0.63 (0.56, 0.69) \\
Inception-V3 & 0.58 (0.52, 0.64) & 0.55 (0.51, 0.59) & 0.49 (0.40, 0.58) & 0.52 (0.44, 0.60) & 0.55 (0.42, 0.68) & 0.55 (0.43, 0.67) & 0.50 (0.40, 0.59) \\
ResNet-50 & 0.56 (0.50, 0.62) & 0.52 (0.48, 0.56) & 0.49 (0.42, 0.57) & 0.49 (0.39, 0.59) & 0.51 (0.38, 0.64) & 0.53 (0.39, 0.67) & 0.46 (0.38, 0.55) \\ \midrule
\multicolumn{8}{c}{Incident (n = 100 fundus images | 50 PD, 50 HC)} \\ \midrule
Logistic Regression & 0.60 (0.54, 0.66) & 0.52 (0.47, 0.56) & 0.53 (0.48, 0.57) & 0.50 (0.44, 0.56) & 0.56 (0.52, 0.61) & 0.47 (0.39, 0.55) & 0.54 (0.50, 0.58) \\
Elastic Net & 0.55 (0.50, 0.61) & 0.52 (0.48, 0.56) & 0.52 (0.48, 0.56) & 0.48 (0.40, 0.56) & 0.63 (0.55, 0.71) & 0.41 (0.32, 0.49) & 0.56 (0.51, 0.61) \\
SVM (Linear) & 0.59 (0.53, 0.65) & 0.53 (0.48, 0.58) & 0.55 (0.50, 0.59) & 0.52 (0.46, 0.58) & 0.58 (0.52, 0.63) & 0.48 (0.40, 0.57) & 0.55 (0.51, 0.59) \\
SVM (RBF) & \textcolor{blue}{0.64 (0.59, 0.70)} & \textcolor{blue}{0.62 (0.58, 0.67)} & \textcolor{blue}{0.63 (0.58, 0.68)} & \textcolor{blue}{0.63 (0.58, 0.67)} & \textcolor{blue}{0.64 (0.59, 0.69)} & \textcolor{blue}{0.60 (0.53, 0.67)} & \textcolor{blue}{0.63 (0.59, 0.67)} \\
Alexnet & \textcolor{red}{0.68 (0.60, 0.75)} & \textcolor{red}{0.60 (0.54, 0.66)} & \textcolor{red}{0.49 (0.39, 0.60)} & \textcolor{red}{0.66 (0.56, 0.77)} & \textcolor{red}{0.64 (0.48, 0.79)} & \textcolor{red}{0.56 (0.45, 0.68)} & \textcolor{red}{0.54 (0.42, 0.67)} \\

VGG-16 & 0.63 (0.54, 0.71) & 0.55 (0.50, 0.60) & 0.54 (0.47, 0.61) & 0.45 (0.31, 0.60) & 0.69 (0.57, 0.82) & 0.40 (0.26, 0.55) & 0.57 (0.50, 0.65) \\
GoogleNet & 0.59 (0.52, 0.66) & 0.56 (0.52, 0.61) & 0.49 (0.41, 0.58) & 0.57 (0.47, 0.67) & 0.63 (0.51, 0.76) & 0.49 (0.38, 0.60) & 0.55 (0.45, 0.64) \\
Inception-V3 & 0.54 (0.47, 0.61) & 0.56 (0.51, 0.61) & 0.56 (0.51, 0.62) & 0.50 (0.39, 0.61) & 0.64 (0.54, 0.75) & 0.48 (0.36, 0.60) & 0.57 (0.51, 0.64) \\
ResNet-50 & 0.58 (0.50, 0.66) & 0.55 (0.52, 0.58) & 0.53 (0.43, 0.63) & 0.42 (0.30, 0.55) & 0.63 (0.48, 0.77) & 0.47 (0.31, 0.64) & 0.52 (0.43, 0.62) \\ \bottomrule
\end{tabular}%
}
\caption{\textbf{Classification Results of Parkinson Disease Datasets}. For each model, the mean average for each performance metric with their 95\% confidence interval is provided over 5 randomized repetitions of 5-fold stratified cross-validation. Highlighted in red represents the best model and highlighted in blue is the best performing conventional machine learning model.}
\label{table: table2}
\end{table}

\newpage
\section*{Supplementary}
\setcounter{table}{0}
\renewcommand{\thetable}{S\arabic{table}}

\begin{table}[h!]
\centering
\begin{tabular}{l l l } 
\toprule
Quartile (diagnostic gap; years) & N & Sensitivity (mean, 95\% CI) \\
\hline
\textbf{Overall Parkinson’s} &	123	 & \\

Q1 (-18.72, -6.32) & 31 & 84.52 (82.64, 85.67) \\ 
Q2 (-6.32, -1.87) & 30 & 84.00 (81.71, 85.32) \\ 
Q3 (-1.87, 4.46) & 30 & 67.00 (72.58, 77.97) \\
Q4 (4.46, 7.38) & 32 & 83.75 (81.40, 85.20) \\ 
\hline
\textbf{Prevalent Parkinson's} & 77 & \\ 

Q1 (-18.72, -7.63) & 19 &  81.05 (77.46, 82.76) \\ 
Q2 (-7.63, -4.80) & 19 & 76.84 (74.14, 78.26) \\ 
Q3 (-4.80, -2.04) & 18 & 83.33 (80.77, 84.57)  \\ 
Q4 (-2.04, 0) & 21 & 78.10 (75.00, 79.67) \\ 
\hline
\textbf{Incident Parkinson’s} & 46 & \\
Q1 (0.09, 3.93) & 11 & 80.00 (73.00, 82.57) \\
Q2 (3.93, 5.07) & 11 & 80.00 (74.11, 82.05) \\
Q3 (5.07, 5.57) & 12 & 93.33 (79.39, 94.83) \\ 
Q4 (5.57, 7.38) & 12  & 81.67 (79.56, 82.47) \\ 
\bottomrule

\end{tabular}
\caption{The performance of the AlexNet model according to each quartile of the Parkinson’s disease diagnostic gap (years) across five randomized repetitions of five-fold cross validation. The diagnostic gap as the difference between the date of image acquisition minus the date of diagnosis, where a negative value is interpreted as having a PD diagnosis prior to fundus image acquisition (prevalent PD) and a positive value is interpreted as having a PD diagnosis post fundus image acquisition (incident PD).}
\label{table: S1}
\end{table}

\begin{table}[h!]
\centering
\begin{tabular}{l l l l}
\toprule
Dataset Category & Gender (M/F) & Metric (Male \% / Female \%) & p-value \\
\hline
\textbf{Overall Parkinson's} \\
PD + HC & 142 / 106 & Accuracy (0.747 / 0.731) & 0.781 \\ 
PD & 71 / 52 & Sensitivity (0.845 / 0.855) & 0.530 \\ 
HC & 71 / 52 & Specificity (0.648, 0.578) & 0.424 \\
\hline
\textbf{Prevalent Parkinson's}  \\
PD + HC & 86 / 68 & Accuracy (0.744 / 0.721) & 0.742 \\ 
PD & 43 / 34 & Sensitivity (0.721 / 0.853) & 0.166 \\ 
HC & 43 / 34 & Specificity (0.767 / 0.588) & 0.092 \\ 
\hline
\textbf{Incident Parkinson's}  \\ 
PD + HC & 56 / 36 & Accuracy (0.768 / 0.694) & 0.434 \\ 
PD & 28 / 18 & Sensitivity (0.929 / 0.848) & 0.093 \\ 
HC & 28 / 18 & Specificity (0.607 / 0.667) & 0.683 \\
\bottomrule
\end{tabular}
\caption{The performance of the AlexNet model according to male and female sub-types. Chi-squared tests of independence were used to test the significance of male and female gender on the model performance. In one instance (Incident PD-specific), a Fischer’s Exact Test as the number of instances in the contingency table was less than or equal to five.}
\label{table: S2}
\end{table}

\newpage
\begin{table}[htbp]
\centering
\resizebox{\textwidth}{!}{
\begin{tabular}{l c c c}
\toprule
Model & Training Time (mean (std), minutes) & Testing Time (mean (std), seconds) & Parameters \\ 
    \midrule
    \multicolumn{4}{c}{Overall (n = 246 fundus images | 123 PD, 123 HC)} \\
    \midrule
    Logistic Regression & 2.73 (0.02) & 0.19 (0.01) & 0.20 \\
    Elastic Net         & 14.01 (0.34) & 0.18 (0.01) & 0.20 \\
    SVM (Linear)        & 2.20 (0.03) & 1.56 (0.01) & 0.20 \\
    SVM (RBF)           & 2.75 (0.03) & 3.57 (0.17) & 0.20 \\
    Alexnet             & 19.13 (2.14) & 7.12 (2.15) & 57.0 \\
    VGG-16              & 28.90 (1.52) & 16.14 (2.12) & 134.3 \\
    GoogleNet           & 18.03 (1.22) & 8.32 (2.43) & 12.0 \\
    Inception-V3        & 21.52 (0.51) & 17.40 (2.12) & 21.8 \\
    ResNet-50           & 20.32 (1.32) & 16.01 (2.50) & 23.5 \\
    \midrule
    \multicolumn{4}{c}{Prevalent (n = 146 fundus images | 73 PD, 73 HC)} \\
    \midrule
    Logistic Regression & 1.86 (0.04) & 0.11 (0.00) & 0.20 \\
    Elastic Net         & 8.45 (0.23) & 0.09 (0.01) & 0.20 \\
    SVM (Linear)        & 0.83 (0.01) & 0.63 (0.04) & 0.20 \\
    SVM (RBF)           & 1.09 (0.01) & 1.55 (0.25) & 0.20 \\
    Alexnet             & 12.20 (0.69) & 9.13 (1.97) & 57.0 \\
    VGG-16              & 16.81 (1.54) & 15.90 (1.96) & 134.3 \\
    GoogleNet           & 11.41 (0.62) & 7.56 (2.06) & 12.0 \\
    Inception-V3        & 14.38 (0.96) & 16.34 (2.70) & 21.8 \\
    ResNet-50           & 14.07 (0.95) & 14.53 (2.18) & 23.5 \\
    \midrule
    \multicolumn{4}{c}{Incident (n = 100 fundus images | 50 PD, 50 HC)} \\
    \midrule
    Logistic Regression & 1.22 (0.06) & 0.07 (0.01) & 0.20 \\
    Elastic Net         & 4.83 (0.22) & 0.08 (0.02) & 0.20 \\
    SVM (Linear)        & 0.26 (0.01) & 0.23 (0.02) & 0.20 \\
    SVM (RBF)           & 0.37 (0.01) & 0.59 (0.05) & 0.20 \\
    Alexnet             & 7.27 (0.43) & 8.48 (1.85) & 57.0 \\
    VGG-16              & 10.13 (0.48) & 15.21 (2.50) & 134.3 \\
    GoogleNet           & 7.03 (0.18) & 7.06 (2.20) & 12.0 \\
    Inception-V3        & 8.36 (0.63) & 16.63 (2.75) & 21.8 \\
    ResNet-50           & 7.11 (0.31) & 14.19 (1.96) & 23.5 \\
    \bottomrule
\end{tabular}}

\caption{The computation time estimates of 5-Fold-Cross-Validation for training and testing of models on color fundus images is recorded by its average and standard deviation. The number of parameters with respect to a $256 \times 256 \times 3$ RGB image input with 2 class outputs is provided for reference and interpretation relative to the model complexity. Estimates were obtained on the University of Florida HiPerGator with 2 CPU cores and 1 NVIDIA A100 GPU.}
\label{table: S3}
\end{table}

\newpage
\begin{table}[htbp]
\centering
\resizebox{\textwidth}{!}{
\begin{tabular}{lccccccc}
\toprule
Model & AUC & ACC & PPV & NPV & SENS & SPEC & F1 \\
\midrule
\multicolumn{8}{c}{Overall (n = 246 fundus images | 123 PD, 123 HC)} \\
\midrule
Logistic Regression (Gray) & 0.69 (0.66, 0.72) & 0.63 (0.61, 0.66) & 0.63 (0.60, 0.65) & 0.65 (0.62, 0.68) & 0.68 (0.64, 0.72) & 0.59 (0.54, 0.63) & 0.65 (0.62, 0.67) \\
Logistic Regression (Vessel) & 0.67 (0.64, 0.70) & 0.61 (0.59, 0.63) & 0.60 (0.57, 0.62) & 0.63 (0.61, 0.65) & 0.69 (0.66, 0.72) & 0.52 (0.47, 0.57) & 0.64 (0.62, 0.66) \\
Elastic Net (Gray) & 0.68 (0.64, 0.71) & 0.63 (0.59, 0.66) & 0.61 (0.58, 0.63) & 0.65 (0.61, 0.70) & 0.72 (0.68, 0.75) & 0.54 (0.49, 0.58) & 0.66 (0.63, 0.68) \\
Elastic Net (Vessel) & 0.67 (0.63, 0.70) & 0.62 (0.59, 0.64) & 0.60 (0.58, 0.62) & 0.65 (0.62, 0.68) & 0.73 (0.69, 0.77) & 0.51 (0.46, 0.56) & 0.65 (0.63, 0.67) \\
SVM (Linear, Gray) & 0.69 (0.66, 0.71) & 0.64 (0.61, 0.66) & 0.63 (0.61, 0.66) & 0.64 (0.62, 0.67) & 0.66 (0.62, 0.70) & 0.61 (0.56, 0.66) & 0.64 (0.62, 0.67) \\
SVM (Linear, Vessel) & 0.68 (0.65, 0.71) & 0.61 (0.59, 0.63) & 0.60 (0.58, 0.62) & 0.62 (0.60, 0.64) & 0.66 (0.62, 0.70) & 0.55 (0.50, 0.60) & 0.63 (0.61, 0.64) \\
SVM (RBF, Gray) & 0.73 (0.70, 0.75) & 0.70 (0.68, 0.72) & 0.67 (0.65, 0.68) & 0.76 (0.72, 0.80) & 0.80 (0.76, 0.84) & 0.59 (0.56, 0.63) & 0.72 (0.70, 0.75) \\
SVM (RBF, Vessel) & 0.65 (0.61, 0.69) & 0.61 (0.58, 0.64) & 0.56 (0.49, 0.64) & 0.64 (0.60, 0.68) & 0.65 (0.56, 0.74) & 0.57 (0.50, 0.64) & 0.60 (0.52, 0.68) \\
\midrule
\multicolumn{8}{c}{Prevalent (n = 146 fundus images | 73 PD, 73 HC)} \\
\midrule
Logistic Regression (Gray) & 0.75 (0.71, 0.78) & 0.67 (0.64, 0.71) & 0.68 (0.64, 0.72) & 0.68 (0.64, 0.72) & 0.69 (0.65, 0.73) & 0.66 (0.60, 0.72) & 0.68 (0.65, 0.71) \\
Logistic Regression (Vessel) & 0.63 (0.58, 0.67) & 0.59 (0.56, 0.63) & 0.58 (0.55, 0.61) & 0.62 (0.57, 0.66) & 0.69 (0.64, 0.73) & 0.50 (0.45, 0.54) & 0.63 (0.59, 0.66) \\
Elastic Net (Gray) & 0.72 (0.68, 0.77) & 0.68 (0.65, 0.71) & 0.66 (0.62, 0.69) & 0.72 (0.69, 0.76) & 0.76 (0.72, 0.81) & 0.59 (0.53, 0.64) & 0.70 (0.68, 0.73) \\
Elastic Net (Vessel) & 0.60 (0.56, 0.65) & 0.57 (0.53, 0.61) & 0.56 (0.52, 0.60) & 0.52 (0.43, 0.62) & 0.70 (0.63, 0.76) & 0.44 (0.35, 0.52) & 0.61 (0.58, 0.65) \\
SVM (Linear, Gray) & 0.74 (0.71, 0.78) & 0.68 (0.64, 0.71) & 0.69 (0.64, 0.73) & 0.69 (0.65, 0.72) & 0.70 (0.65, 0.74) & 0.66 (0.60, 0.73) & 0.68 (0.65, 0.72) \\
SVM (Linear, Vessel) & 0.63 (0.59, 0.67) & 0.57 (0.53, 0.61) & 0.56 (0.53, 0.60) & 0.58 (0.54, 0.62) & 0.63 (0.58, 0.68) & 0.51 (0.46, 0.56) & 0.59 (0.55, 0.63) \\
SVM (RBF, Gray) & 0.74 (0.70, 0.78) & 0.72 (0.69, 0.75) & 0.69 (0.66, 0.72) & 0.79 (0.75, 0.84) & 0.83 (0.78, 0.87) & 0.61 (0.56, 0.66) & 0.75 (0.72, 0.77) \\
SVM (RBF, Vessel) & 0.58 (0.53, 0.64) & 0.58 (0.54, 0.62) & 0.52 (0.44, 0.60) & 0.59 (0.52, 0.66) & 0.62 (0.50, 0.74) & 0.54 (0.45, 0.63) & 0.56 (0.46, 0.65) \\
\midrule
\multicolumn{8}{c}{Incident (n = 100 fundus images | 50 PD, 50 HC)} \\
\midrule
Logistic Regression (Gray) & 0.56 (0.50, 0.61) & 0.53 (0.49, 0.56) & 0.53 (0.49, 0.57) & 0.52 (0.48, 0.57) & 0.56 (0.50, 0.63) & 0.49 (0.42, 0.56) & 0.54 (0.49, 0.58) \\
Logistic Regression (Vessel) & 0.74 (0.69, 0.79) & 0.65 (0.60, 0.71) & 0.66 (0.60, 0.72) & 0.68 (0.61, 0.74) & 0.69 (0.62, 0.76) & 0.62 (0.53, 0.71) & 0.66 (0.61, 0.72) \\
Elastic Net (Gray) & 0.55 (0.51, 0.60) & 0.54 (0.51, 0.58) & 0.52 (0.46, 0.57) & 0.53 (0.44, 0.61) & 0.65 (0.55, 0.75) & 0.44 (0.34, 0.53) & 0.57 (0.50, 0.63) \\
Elastic Net (Vessel) & 0.74 (0.69, 0.78) & 0.65 (0.61, 0.70) & 0.66 (0.61, 0.71) & 0.68 (0.62, 0.74) & 0.68 (0.60, 0.75) & 0.63 (0.56, 0.71) & 0.65 (0.60, 0.70) \\
SVM (Linear, Gray) & 0.54 (0.48, 0.60) & 0.51 (0.48, 0.55) & 0.52 (0.48, 0.56) & 0.51 (0.47, 0.54) & 0.49 (0.44, 0.55) & 0.53 (0.46, 0.60) & 0.50 (0.46, 0.54) \\
SVM (Linear, Vessel) & 0.75 (0.70, 0.80) & 0.66 (0.61, 0.71) & 0.67 (0.62, 0.72) & 0.69 (0.62, 0.76) & 0.68 (0.60, 0.75) & 0.64 (0.56, 0.72) & 0.66 (0.61, 0.71) \\
SVM (RBF, Gray) & 0.66 (0.60, 0.72) & 0.63 (0.58, 0.69) & 0.58 (0.49, 0.68) & 0.65 (0.57, 0.73) & 0.65 (0.53, 0.76) & 0.62 (0.54, 0.70) & 0.60 (0.50, 0.70) \\
SVM (RBF, Vessel) & 0.56 (0.46, 0.66) & 0.57 (0.54, 0.61) & 0.40 (0.27, 0.54) & 0.58 (0.50, 0.66) & 0.43 (0.28, 0.59) & 0.72 (0.59, 0.84) & 0.40 (0.27, 0.53) \\
\bottomrule
\end{tabular}}
\caption{Gray-Scale Conversion and Vessel Segmentation Classification Results of Machine Learning Models. Values presented as mean, 95\% confidence interval.}
\label{table: S4}%
\end{table}%

\end{document}